\pdfoutput=1

\documentclass[11pt]{article}

\usepackage[]{EMNLP2022}

\usepackage{times}
\usepackage{latexsym}
\usepackage{placeins}

\usepackage[T1]{fontenc}

\usepackage[utf8]{inputenc}

\usepackage{microtype}

\usepackage{inconsolata}

\usepackage{times}
\usepackage{latexsym}
\usepackage{graphicx}
\usepackage{booktabs}
\usepackage{multirow}
\usepackage{float}
\usepackage{fixltx2e}
\usepackage[LGRgreek]{mathastext}

\usepackage[T1]{fontenc}
\usepackage{amsmath}

\usepackage[utf8]{inputenc}

\usepackage{microtype}
\usepackage{dsfont}
\usepackage{tabularx}

\newcommand\ourmodel{kNN-Prompt}

\DeclareSymbolFont{matha}{OML}{txmi}{m}{it}
\DeclareMathSymbol{\varv}{\mathord}{matha}{118}

\title{kNN-Prompt: Nearest Neighbor Zero-Shot Inference}

\author{Weijia Shi  \quad
Julian Michael \quad
Suchin Gururangan  \quad Luke Zettlemoyer\\
  Paul G. Allen School of Computer Science \& Engineering, \\ University of Washington, Seattle, WA \\
  \texttt{\{swj0419, julianjm, sg01, lsz\}@cs.washington.edu}
  }
  
\begin{document}
\maketitle

\maketitle
\begin{abstract}
 
Retrieval-augmented language models (LMs) use non-parametric memory
to substantially outperform their non-retrieval counterparts on perplexity-based evaluations, but it is an open question whether they achieve similar gains in few- and zero-shot end-task accuracy. 
We extensively study one such model, the $k$-nearest neighbor LM ($k$NN-LM), showing that the gains marginally transfer. The main challenge is to achieve coverage of the verbalizer tokens that define the different end-task class labels. 
To address this challenge, we also introduce \ourmodel, a simple and effective $k$NN-LM with automatically expanded fuzzy verbalizers (e.g. to expand ``terrible'' to also include ``silly'' and other task-specific synonyms for sentiment classification). 
Across nine diverse end-tasks, using \ourmodel~with GPT-2 large yields significant performance boosts over strong zero-shot baselines (13.4\% absolute improvement over the base LM on average).
We also show that other advantages of non-parametric augmentation hold for end tasks; \ourmodel\ is effective for domain adaptation with no further training, and gains increase with the size of the retrieval model.

\end{abstract}
\section{Introduction}

\begin{figure}[t]
\centering
    \includegraphics[scale=0.34]{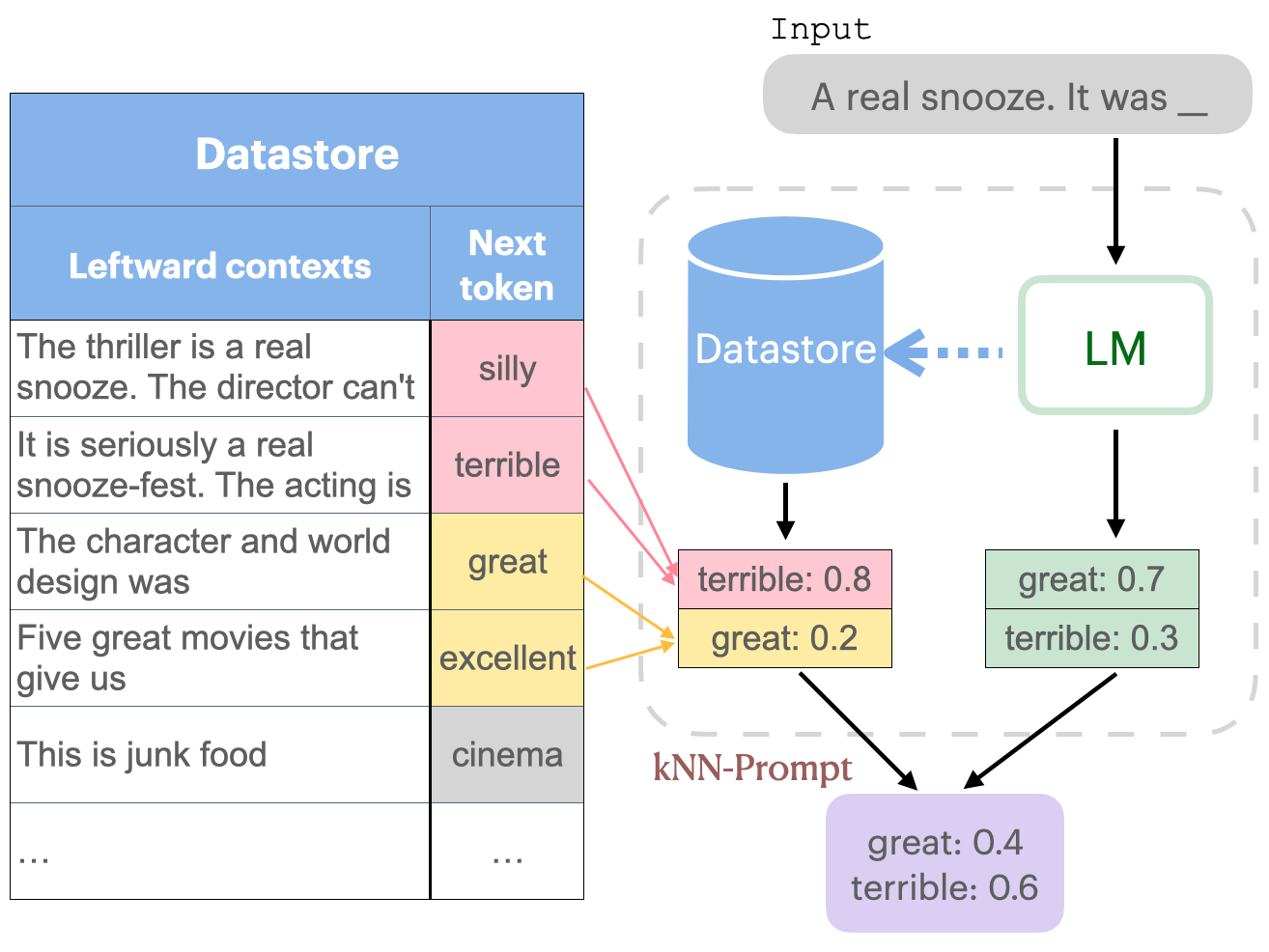}
    \caption{
    \ourmodel~incorporates information from a large, heterogeneous corpus (unlabeled texts from different domains) to facilitate few- and zero-shot inference. The datastore contains key-value pairs where the key is an encoding of a leftward context and the value is the next token following the context. Our fuzzy verbalizer expands "terrible" to include "silly" and "great" to include "excellent". Because the encoded corpus is unlabeled plain text, some datastore entries contain next tokens not in the verbalizer tokens (e.g., "cinema").  
    }
    \label{fig:first}
\end{figure}

Retrieval-augmented language models (LMs) have access to a non-parametric memory, allowing them to directly access a large external text collection during inference. Previous work has shown that these models substantially outperform their non--retrieval-based counterparts on language modeling tasks~\cite{Khandelwal2020Generalization, he2021efficient, borgeaud2021improving}, but it is an open question whether they also achieve similar gains in few-shot and zero-shot end task evaluations~\cite{radford2019language, brown2020language}.  In this paper, we demonstrate that, with some extensions to improve coverage of the verbalizer tokens, the performance gains of retrieval-augmented LMs generalize well to a wide range of downstream tasks.
%




We study the $k$-nearest neighbors language model \cite[$k$NN-LM]{Khandelwal2020Generalization}, which interpolates the LM softmax distribution with a nearest-neighbor distribution.
The nearest neighbours are computed based on the distance in LM output embeddings and can be drawn from any text corpus, in our case, a heterogeneous corpus that contains unlabeled data from different domains. 
We are the first to study the zero-shot application of $k$NN-LM to end tasks, and 
 we find that applying the technique na\"{i}vely produces only marginal improvements (\autoref{sec:results}).
The main challenge is that the support of the $k$NN distribution is sparse (covering at most $k$ tokens, often less), as it only assigns probability mass to nearest neighbors.
This means it often entirely misses the tokens that are used to verbalize the output label in the standard application of LMs to zero-shot classification: across the datasets we test, an output label receives nonzero probability under the $k$NN distribution only 44.2\% of the time (see \autoref{sec:ablation}).

To address this challenge, we introduce \ourmodel, a simple and effective method built on $k$NN-LM for improving zero-shot inference with no further training.
Key to our approach are \textit{fuzzy verbalizers}, which automatically expand the set of tokens corresponding to each output label.
For example, in \autoref{fig:first}, the verbalized label of the negative sentiment is ``terrible.'' Our fuzzy verbalizer also maps ``silly'' to negative sentiment, allowing the model to better leverage the information available in the $k$NN distribution.
Extensive experiments (\autoref{sec:experimental-setup}) show that applying \ourmodel\ using a purely unlabeled heterogeneous corpus consistently improves zero-shot performance on eleven tasks, including sentiment analysis, topic classification, entailment, fact retrieval and question answering. These improvements hold for every model in the GPT-2 family.

We also show that \ourmodel\ can be used to adapt LMs to new domains and tasks with no further training (\autoref{sec:adaptation}).
With a domain-specific datastore corpus, we achieve comparable or better performance to prompting the LM after domain-adaptive pretraining \citep{gururangan-etal-2020-dont} on that corpus.
To better understand these gains, we conduct a thorough analysis (\autoref{sec:ablation}), showing that fuzzy verbalizers are essential for leveraging the $k$NN distribution, the benefits of retrieval increase with retrieval model size, and even relatively small datastores can yield sizeable performance gains if they are tailored to the domain or task. Overall, our results show how retrieval can benefit zero-shot inference with LMs on a wide variety of tasks, and suggest that applying retrieval with larger models may yield even greater benefits. Code is available at \url{github.com/swj0419/kNN_prompt}.



%
%

\begin{figure*}
    \centering
    \includegraphics[scale=0.6]{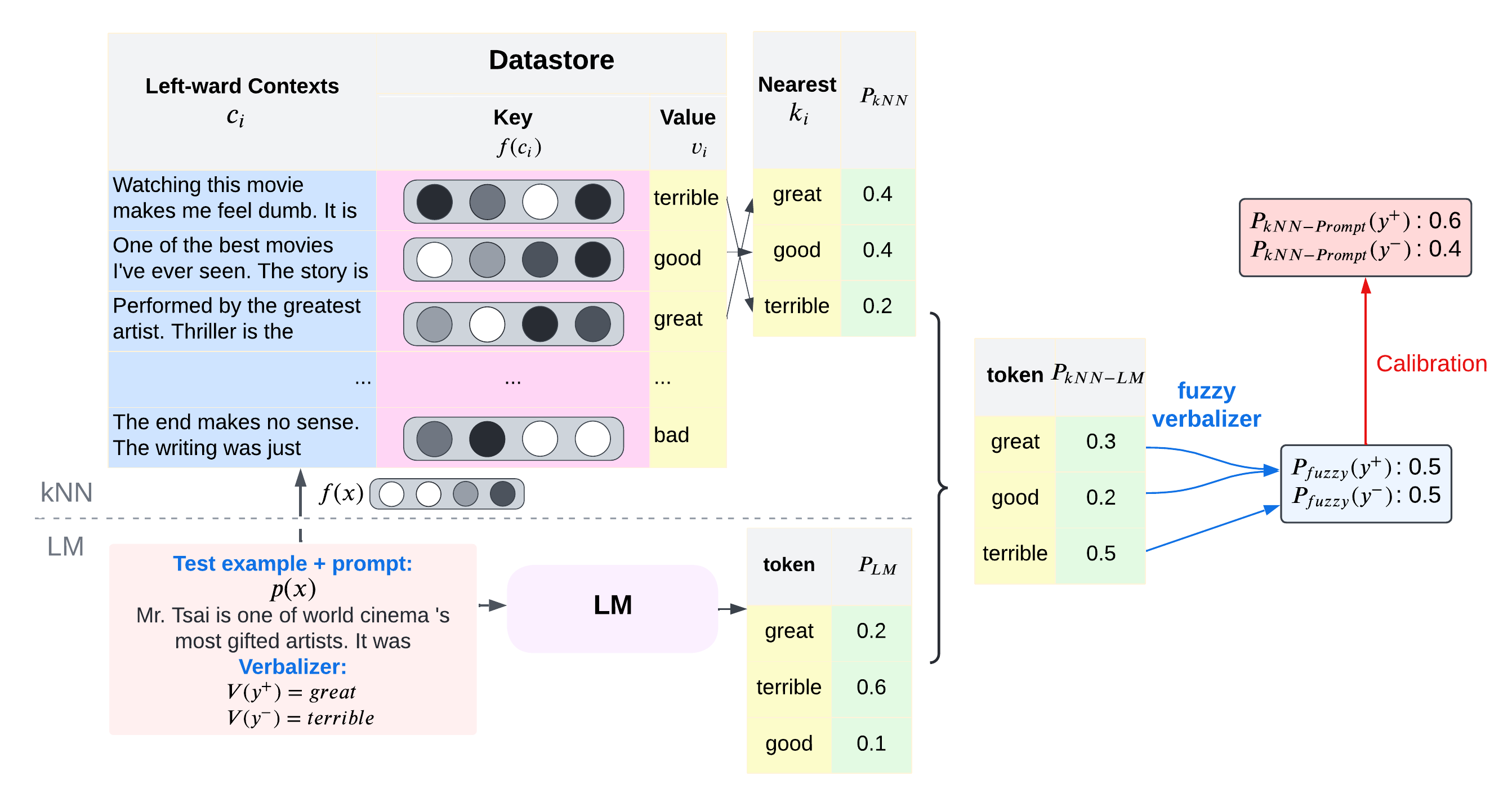}
    \caption{An illustration of \ourmodel~applying to sentiment analysis tasks. Texts are encoded in the datastore, where each entry consists of a representation of a leftward context and its next token. During inference, a test example is mapped to a prompt form and used to retrieve the k most similar contexts and their next tokens from the datastore. The kNN distribution is a multinomial computed on the distance of the text example and similar contexts. The final prediction is formed by combining the kNN distribution with the language model output distribution. 
    }
    \label{fig:knn_prompt}
\end{figure*}



%

\section{Method} \label{sec:method}


To perform zero-shot prediction on a downstream task using a pretrained language model, we recast the task as language modeling \citep{radford2019language} by converting each input instance into a natural language prompt (\autoref{sec:prompting}).
We then augment the pretrained model with the $k$-nearest-neighbors language modeling technique from \citet{Khandelwal2020Generalization}.
To better benefit from the sparse $k$NN distribution, we introduce \textit{fuzzy verbalizers} for mapping from the LM's outputs to a distribution over task-specific labels (\autoref{sec:fuzzy-verbalizers}).
Finally, we decode the output from this label distribution using the domain-conditional PMI scoring method of \citet{holtzman-etal-2021-surface}.

\subsection{Prompting and Verbalizers} 
\label{sec:prompting}
We address classification problems
where an instance consists of an input sequence of tokens $\mathbf{x} = (x_0, x_1, ..., x_{|\mathbf{x}|})$ from a vocabulary $\mathcal{V}$ and an output label $y \in Y$. 
The output label set $Y$ may be fixed for the task (\textit{text classification}).
For example, in the sentiment analysis example in \autoref{fig:knn_prompt}, the input is $\mathbf{x} = $ ``Mr.~Tsai is one of world cinema's most gifted artists.'' The output labels are $Y$ = \{$y^+$, $y^-$\}, referring to positive and negative sentiment. 

To cast the task as language modeling, we deterministically transform each input example $\mathbf{x}$ into a \textbf{prompt} $p(\mathbf{x})$.
Providing this prompt to an LM yields a probability distribution
$ P_{LM}(\mathbf{v} \mid p(\mathbf{x})) $.
To extract an output label from this, we apply \textit{verbalizers} $V: y \rightarrow
\mathcal{V}^*$ \cite{schick-schutze-2021-exploiting} which map each output label $y \in Y$ to a label word $V(y) = \mathbf{v}$.
We can then compute a probability for each label:
\begin{equation} \label{eq:prompt-label-prob}
P(y \mid \mathbf{x}) \propto P_{LM}(V(y) \mid p(\mathbf{x})),
\end{equation}
normalizing over all $y \in Y$.

For example, our prompt transformation for sentiment analysis adds \textit{It was} after the input, and uses the verbalizer $V(y^+) = \textit{great}, V(y^-) = \textit{terrible}$, which classifies sentiment according to the relative probabilities of \textit{It was great} and \textit{It was terrible} after the input sequence (see \autoref{fig:knn_prompt}, bottom left).

\subsection{$k$-Nearest Neighbors Language Modeling}
\label{sec:knnlm}

Following  \citet{Khandelwal2020Generalization}, we
augment the LM with a \textit{datastore} from which it can retrieve tokens that inform its predictions, improving performance without further training.

The datastore is a key-value store generated by running the LM over a corpus of text.
Each value is a token $w \in \mathcal{V}$ from the corpus, and its key is the vector hidden representation at the output layer of the LM running forward on the left context $\mathbf{c} \in \mathcal{V}^*$ (call this $f(\mathbf{c})$).
At inference time, when predicting the next token for an input sequence $\mathbf{c}$, the $k$NN-LM retrieves the $k$ nearest neighbors of $\mathbf{c}$ from the datastore according to the distance $d(\cdot, f(\mathbf{c}))$ of their key vectors (squared $L^2$ distance following \citeauthor{Khandelwal2020Generalization}). 

A softmax over the (negative) distances induces a distribution over the the tokens $w_i$ in the nearest neighbor set:
 $$P_{k\text{NN}} (v \mid \mathbf{c})  \propto   \sum_{(f(\mathbf{c}_i), w_i)} \mathds{1}_{v = w_i} e^{\frac{- d( f(\mathbf{c}_i), f(\mathbf{c}) )}{t}}$$
where $t$ is a temperature parameter.\footnote{We have added the temperature adjustment in the softmax on top of the $k$NN-LM formulation.}
We can then interpolate this with the original LM as follows:
\begin{equation*} \label{eq:knnlm}
P_{k\text{NN-LM}}(v \mid c) =
  (1-\lambda) P_\text{LM}(v | c)
  + \lambda P_{k\text{NN}} (v | c).
\end{equation*}
The hyperparameters for the $k$NN-LM approach are the number $k$ of nearest neighbors, the interpolation constant $\lambda$, the temperature $t$, and the choice of datastore.

\subsection{Fuzzy verbalizers}
\label{sec:fuzzy-verbalizers}
One challenge in performing zero-shot inference with LMs on downstream tasks is the choice of verbalizer.
On one hand, LMs may be highly sensitive to the particular surface form in ways that are irrelevant to the classification task \citep{holtzman-etal-2021-surface}.
On the other hand, for a $k$NN model, the $k$ nearest neighbor set is sparse and may fail to cover any of the tokens in the set of verbalizers (i.e., $P_{kNN}(V(y)) = 0$ for all $y \in Y$), limiting its utility in those cases.
To address these issues, we introduce \textit{fuzzy verbalizers}, which associate each label $y$ with a neighborhood of token sequences around a specific verbalization $V(y) \in \mathcal{V}^*$. 

To do this, we first associate each token $v \in \mathcal{V}$ with a neighborhood $\mathcal{N}(v) \subseteq \mathcal{V}$ of similar tokens.
In particular, we use $v$'s top-5 most similar words according to the cosine similarity of their GloVe embeddings \citep{pennington-etal-2014-glove}, as well as any of $v$'s synonyms in ConceptNet~\citep{speer2017conceptnet}.\footnote{\url{https://conceptnet.io}}
Then, for the purposes of calculating the probability of a verbalized label $\mathbf{v} = V(y)$, we treat a prediction of any token in each neighborhood of $v$ as a viable substitute for it, marginalizing over $\mathcal{N}(z)$:
\begin{equation} \label{eq:fuzzy}
P_{FV}(y \mid x) \propto \sum_{v_i \in \mathcal{N}(v)} P(v_i \mid p(\mathbf{x}))
\end{equation}
The fuzzy verbalizer helps mitigate the effect the sparsity of the $k$NN distribution has on zero-shot prediction (see \autoref{aba:pmi_vs_fuzzy}).

\subsection{Full model}
\label{sec:full-model}

To make a zero-shot prediction for an input $\mathbf{x}$, we first transform it into a prompt $p(\mathbf{x})$ and obtain a distribution over the label word $\mathbf{v}$ with a $k$NN-LM:
$P_{kNN-LM}(\mathbf{v} \mid p(\mathbf{x}))$.
We then apply \textit{domain-conditional PMI} scoring rule \citep{holtzman-etal-2021-surface} to calibrate the distribution: 
$$PMI_{DC}(v, p(\mathbf{x})) = \frac{P(v \mid p(\mathbf{x}))}{P(v \mid \mathbf{p})}$$
where $\mathbf{p}$ is a task-dependent string which is independent of the particular input (generally the local context at the end of the prompt, e.g., we use $\mathbf{p} = \text{``It was''}$ for sentiment analysis, as shown in \autoref{fig:knn_prompt}).

Finally, we convert this to the output label score $P(y \mid p(\mathbf{x}))$ using a fuzzy verbalizer (\autoref{sec:fuzzy-verbalizers}). When using the fuzzy verbalizer together with PMI calibration, instead of marginalizing over the tokens in the fuzzy verbalizer $v_i \in \mathcal{N}(v)$ (\autoref{eq:fuzzy}), we score each label according to the sum of the PMIs of its associated tokens:

$$
P(y \mid \mathbf{x}) \propto  \sum_{v_i \in \mathcal{N}(v)} PMI_{DC}(v_i, p(\mathbf{x}))
$$

\section{Experimental Setup}
\label{sec:experimental-setup}

\begin{table}
\small
\centering
\begin{tabular}{lll}
\toprule
\bf Corpus & \bf Size & \bf \# Tokens \\
\midrule
Wikitext-103 & 181MB & 114M \\
Amazon Reviews & 89MB & 19M \\
CC-NEWS & 457MB & 324M \\
IMDB & 45MB & 8M \\
\midrule
Total & 722MB & 465M \\
\bottomrule
\end{tabular}
\caption{
Statistics of our heterogeneous datastore corpora.
} \label{data_info}
\end{table}

\subsection{Tasks} 
We experiment with 9 tasks, including 
topic classification, sentiment analysis, entailment and partisanship classification. 

\paragraph{Topic Classification}
We use the AG News (\textbf{AGN}) and Yahoo!\,Answers (\textbf{Yahoo}) corpora from \citet{zhang2015character} for topic classification.

\paragraph{Sentiment and Partisanship}
We study sentiment analysis using the Rotten Tomatoes (\textbf{RT}) and \textbf{SST-2} corpora of \citet{socher2013recursive}, movie reviews from \citet[\textbf{MR}]{pang-lee-2005-seeing}, the customer review dataset from \citet[\textbf{CR}]{10.1145/1014052.1014073} consisting of Amazon and Yelp reviews, and 
the hyperpartisan news detection dataset from \citet[\textbf{HYP}]{kiesel-etal-2019-semeval}, which focuses on classifying whether a text exhibits extreme political views.

\begin{table*}[htbp]
\centering
\small
\begin{tabular}{lcccccccccc}
\toprule

& RTE & CB & Yahoo & RT & SST-2 & CR & MR & HYP & AGN & Avg \\
\midrule
LM & 53.1 & 48.2 & 49.7 & 53.0 & 55.3 & 66.2 & 54.6 & 58.5 & 67.4 & 56.2 \\
LM+PMI & 54.2 & 50.0 & 48.8 & 74.1 & 76.5 & 82.8 & 74.6 & 58.5 & 65.1 & 65.0 \\
kNN-LM & 53.1 & 48.2 & 49.5 & 54.5 & 55.4 & 67.2 & 56.4 & 58.5 & 67.0 & 56.6 \\
kNN-Prompt & \bf 55.6 & \bf 53.5 & \bf 51.0 & \bf 80.6 & \bf 84.2 & \bf 84.3 & \bf 78.2 & \bf 60.0 & \bf 78.8 & \bf 69.6 \\
\bottomrule

\end{tabular}

\caption{
Zero-shot results on a variety of tasks.
Our model, \ourmodel, handily outperforms \citet{holtzman-etal-2021-surface}'s PMI scoring method alone (LM+PMI) as well as the base $k$NN-LM method of \citet{Khandelwal2020Generalization}.
} 
\label{main_table}

\end{table*}






\paragraph{Entailment}
Entailment datasets focus on classifying whether one sentence plausibly implies another to be true or false.
We evaluate on the CommitmentBank \citep[\textbf{CB}]{de2019commitmentbank} and the Recognizing Textual Entailment~\citep[\textbf{RTE}]{dagan_dolan_magnini_roth_2010} dataset provided in GLUE~\cite{wang-etal-2018-glue}.

\subsection{\ourmodel\ Model Details}
\label{sec:knnpromptdetails}
\paragraph{Inference Model}
For our main experiments, we directly use GPT-2 large from Huggingface\footnote{\url{https://github.com/huggingface/transformers}} as our base LM. We consider other model sizes in \autoref{sec:ablation}.

\paragraph{Retriever Model}
Following the inference model, we use GPT-2 large to build the datastore.
The keys are the 1280-dimensional hidden representations before the final MLP which predicts the token distribution at each timestep, produced using a single forward pass over the datastore corpus. 
For efficient similarity search, we create a FAISS~\cite{johnson2019billion} index and search for nearest neighbors by Euclidean distance.  


\paragraph{Datastore Corpus}
For our datastore, we aim to curate a large, heteregenous corpus of data broadly relevant to the tasks we evaluate. To this end, we combine four sources of data including Wikitext-103~\cite{merity2016pointer}, the Amazon review corpus of \citet{he2016ups}, and subsets of CC-NEWS\footnote{\url{https://huggingface.co/datasets/cc\_news}} and IMDB\footnote{\url{http://ai.stanford.edu/~amaas/data/sentiment}} sampled uniformly from each. 
\autoref{data_info} lists the specifics of each data source. 



\paragraph{Inference Procedure} 
We retrieve k=1024 neighbors, soften the kNN distribution with a temperature value of 3 and use an interpolation factor of $\lambda = 0.3$. Our primary evaluation is zero-shot.
All hyperparameters were chosen on the basis of development experiments (see  \autoref{sec:neighbor} for more details).


\subsection{Baselines}

\paragraph{LM} is the result of prompting the base language model (GPT-2 Large), choosing the output label whose verbalizer has the highest probability under the language model $P_{LM}(V(y) \mid p(\mathbf{x}))$. 

\paragraph{LM+PMI} is the approach of 
\citet{holtzman-etal-2021-surface}, calibrating \textbf{LM} with domain-conditional PMI scoring (\autoref{sec:full-model}). 



\paragraph{$k$NN-LM} directly applies the $k$NN-LM of \citet{Khandelwal2020Generalization} in the same way as \textbf{LM}, choosing the highest-probability output label.



\begin{table}[t!]
\centering
\small
\begin{tabular}{lccc}
\toprule
& \bf CR & \bf HYP & \bf MR \\
\midrule
LM         & 79.5 \textsubscript{4.1}        &   56.7 \textsubscript{0.5}       & 78.2 \textsubscript{1.4} \\
LM+PMI     & 79.8 \textsubscript{5.5}         & 52.7 \textsubscript{2.6}          & 76.3 \textsubscript{1.5} \\
$k$NN-LM   & 79.5 \textsubscript{4.2}         & 56.7 \textsubscript{1.5}         & 77.5 \textsubscript{2.3} \\
kNN-prompt & \bf 80.5 \textsubscript{1.7} & \bf 57.1 \textsubscript{1.1} & \bf 79.4 \textsubscript{1.5} \\
\bottomrule
\end{tabular}
\caption{
The mean and standard deviation for 4 uniformly sampled sets of 4 demonstration examples used for few-shot inference. 
}
  \label{tab:few-shot}

\end{table}

\section{Experimental Results}
\label{sec:results}

Results for zero-shot
prediction are in 
\autoref{main_table}.
\ourmodel\ outperforms all baselines in all tasks, improving over the base LM by 13.4\% on average.
The gains are particularly pronounced for MR and RT (sentiment analysis on movie reviews), Yahoo (topic classification).
For MR and RT,
the gains seem to come mostly from PMI calibration. 


Interestingly, the $k$NN-LM alone yields a fairly small improvement over the base LM (about 0.4\% on average).
This suggests that the fuzzy verbalizer and PMI calibration methods may help \ourmodel\ better leverage the information in the $k$-nearest neighbors distribution.
We examine possible sources of \ourmodel's performance gains  in \autoref{sec:ablation}.

\paragraph{Few-shot inference}
For a subset of tasks, we additionally compare to a few-shot setting where we prepend four examples uniformly sampled from the training data to the input (\autoref{tab:few-shot}). The demonstration examples are converted to prompt and verbalizer format. 
We report the mean accuracy and standard deviation with 4 different random seeds. We find that \ourmodel~consistently outperform baselines, demonstrating that \ourmodel~is applicable to the few-shot setting as well. We leave further exploration of this phenomenon to future work.

\begin{figure*}[ht!]
    \centering
    \includegraphics[width=0.85\textwidth, scale=0.3]{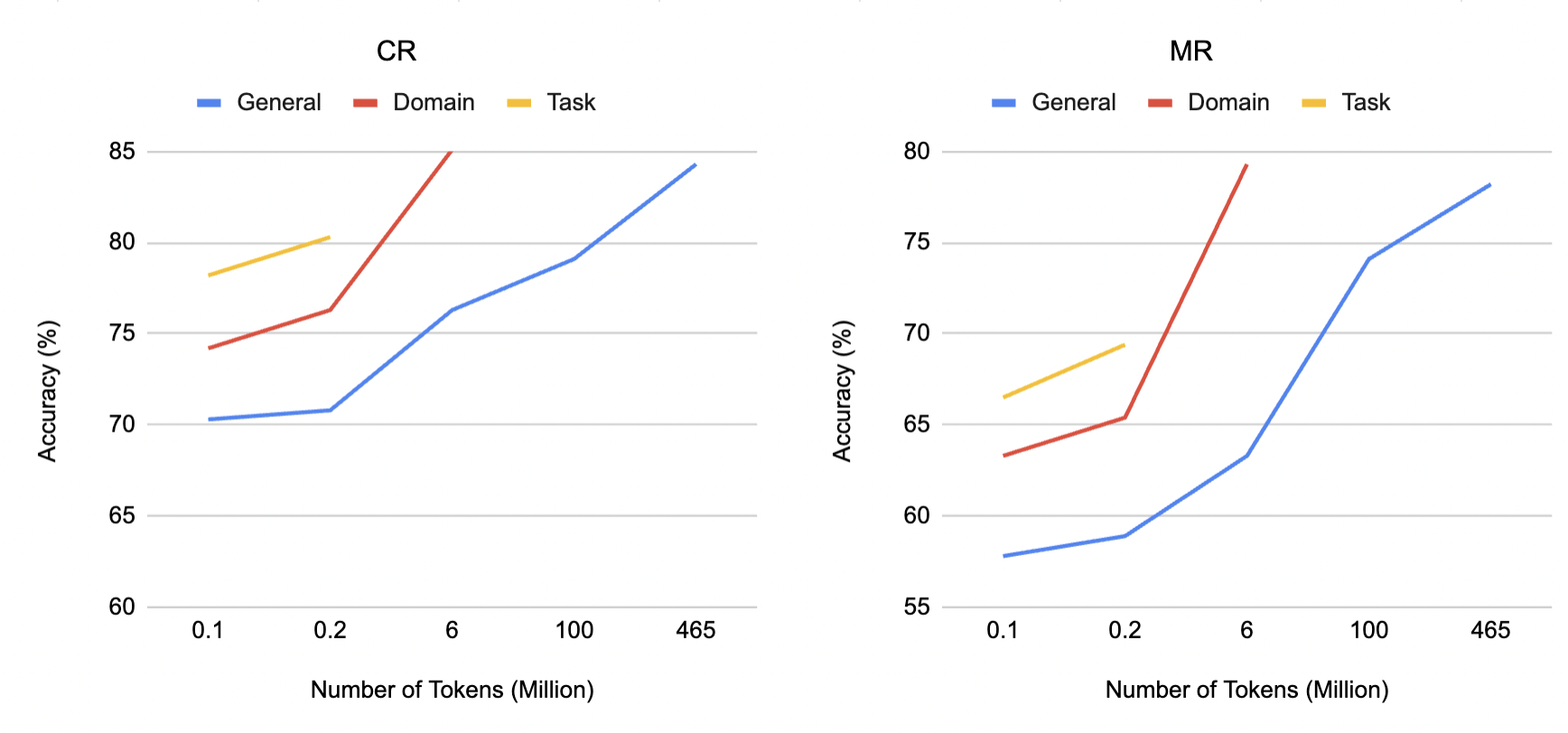}
    \caption{Effect of the number of tokens in the datastore on CR and MR. Each line represents the \ourmodel\ model with a different datastore and the line ends when the entire available datastore is used. 
    General, Domain, and Task refer to the heterogeneous corpus (\autoref{data_info}), domain-specific corpus, and task-specific corpus, respectively. 
    We use IMDB as the domain-specific corpus for MR, and Amazon Reviews for CR. The task-specific corpus is the unlabeled training data of each task. GPT-2 Large is used for both retriever and inference models.
    }
    \label{fig:heatmap}
\end{figure*}

\section{\ourmodel\ for Domain Adaptation} \label{sec:adaptation}
One of the advantages of retrieval-based LMs is that they can be adapted to new domains with no further training.

To test this capability, we replace our heterogeneous datastore (Section \ref{sec:knnpromptdetails}) with domain-specific ones for several tasks. To build these domain-specific datastores, we select Amazon Reviews for CR, CC-NEWS for HYP and IMDB for MR, and encode them separately. 
\begin{table}[t!]
\centering
\small
\begin{tabular}{llccc}
\toprule
& \bf CR & \bf HYP & \bf MR \\
\midrule
LM + PMI & 82.8 & 58.5 & 74.6 \\
kNN-prompt & \bf 84.3 & 60.0 & \bf 78.2 \\
DAPT (LM + PMI) & 84.1 & \bf 61.1 & 77.8\\
\bottomrule
\end{tabular}
\caption{Domain adaptation experiments using domain-specific datastores. DAPT requires training the LM on the corresponding datastore, while \ourmodel\ can use it as the datastore with no further training.
}  \label{adap_table}

\end{table}

For comparison, we consider domain-adaptive pretraining \citep[DAPT]{gururangan-etal-2020-dont}, which further trains the LM on the domain-specific corpus. We train GPT-2 Large on each domain corpus for a single pass, then apply it to downstream tasks using our prompting and verbalizer setup and domain-conditional PMI scoring.

As shown in \autoref{adap_table}, \ourmodel\ performs comparably with DAPT. Specifically, \ourmodel\ slightly outperforms DAPT on CR and MR.
These results indicate that \ourmodel\ is an effective method for domain adaptation. Critically, unlike DAPT, \ourmodel\ does not require further training, which is more practical and efficient for adapting very large LMs. 

 
\paragraph{Effect of datastore distribution and size}
To better understand \ourmodel's potential for domain adaptation, we experiment with varying sizes and distributions of the datastore.
For each task, we consider three options for the datastore corpus: our heterogeneous corpus (Section \ref{sec:knnpromptdetails}), a domain-specific corpus, and a task-specific corpus constructed from the task's (unlabeled) training data. Each of these data sources exhibits increasing levels of relevance to the task.

\autoref{fig:heatmap} shows how model performance varies with the choice of datastore across different datastore sizes.
For a fixed number of tokens, retrieving from a task-specific datastore is best. Furthermore, token-for-token, adding task-specific data leads to more gains than domain-specific data, which in turn is better than our heterogeneous corpus.

 Using domain-specific data is always better than retrieving from the large heterogeneous corpus.
For example, for CR, using 6M tokens of domain-specific data outperforms using our 465M token heterogeneous corpus.
These results suggest that while having a large datastore is beneficial, curating task-specific or domain-specific data can also be an effective way of improving model performance, especially if datastore size is limited (e.g., due to memory constraints).

\begin{table}[]
\small
\begin{tabular}{lrr}
\toprule
\bf Model & \bf Acc. & \bf $\Delta$Acc. \\
\midrule
LM & 56.2 & 0\\
LM+kNN (kNN-LM) & 56.6 & +0.4 \\
LM+Fuzzy & 63.4 & +7.2 \\
LM+PMI & 65.0 & +8.8 \\
\midrule
LM+Fuzzy+PMI & 67.1 & +10.9\\
LM+kNN+Fuzzy & 66.5 & +10.3 \\
LM+kNN+PMI & 64.2 & +8.0 \\
\midrule
LM+kNN+Fuzzy+PMI (kNN-Prompt) & \bf 69.6 & \bf +13.4 \\
\bottomrule
\end{tabular}
\caption{
Effect of different components on the average zero-shot accuracy across the eleven tasks.
} \label{tbl:ablation}
\end{table}

\section{Analysis}
\label{sec:ablation}

\begin{figure*}[h!]
    \centering
    \includegraphics[width=0.85\textwidth, scale=0.3]{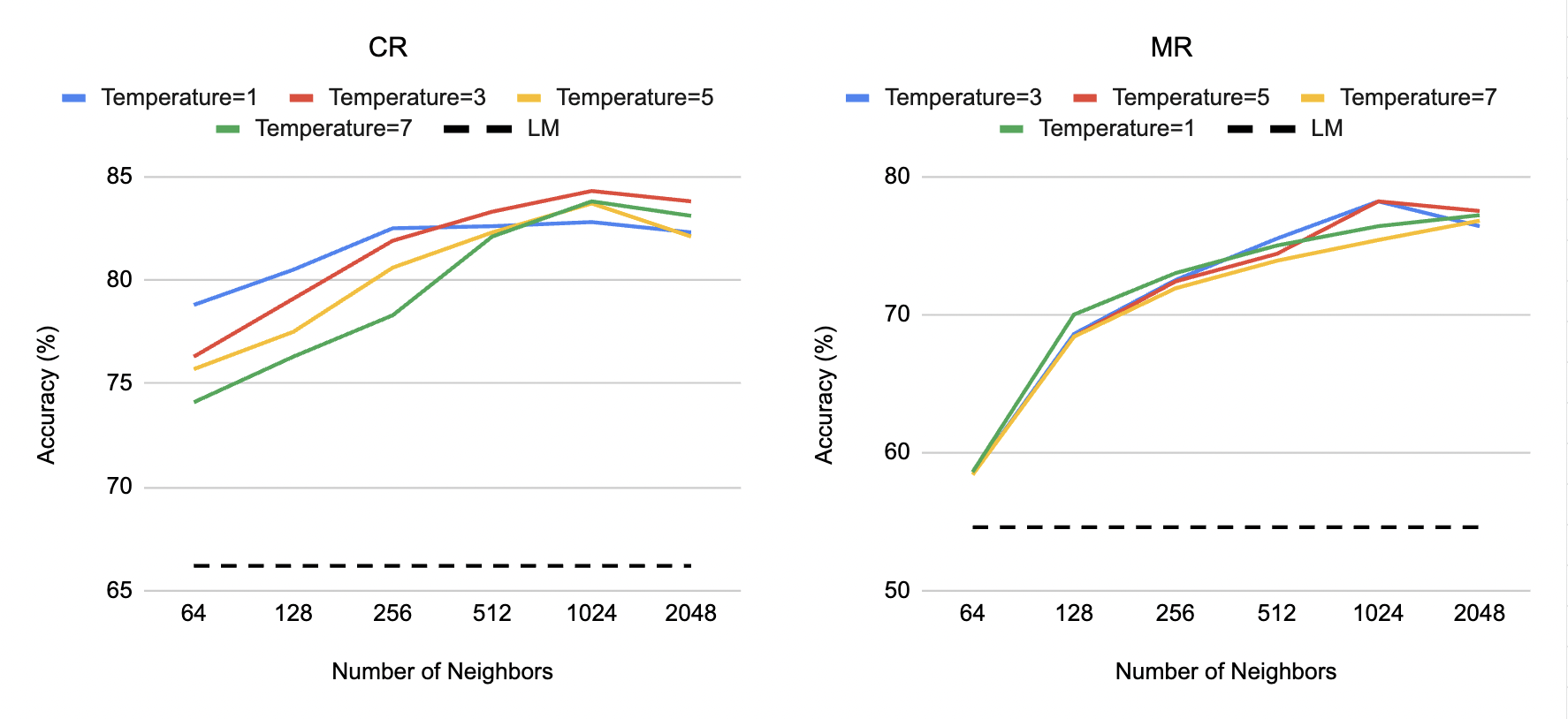}
    \caption{Effect of the number of retrieved neighbors and softmax temperature on \ourmodel\ performance for three tasks: CR and MR. Task performance monotonically improves with the number of neighbors as $k$ is increased to 1024.
    }
    \label{fig:neighbor}
\end{figure*}

\begin{figure*}[h!]
    \centering
    \includegraphics[width=0.85\textwidth, scale=0.3]{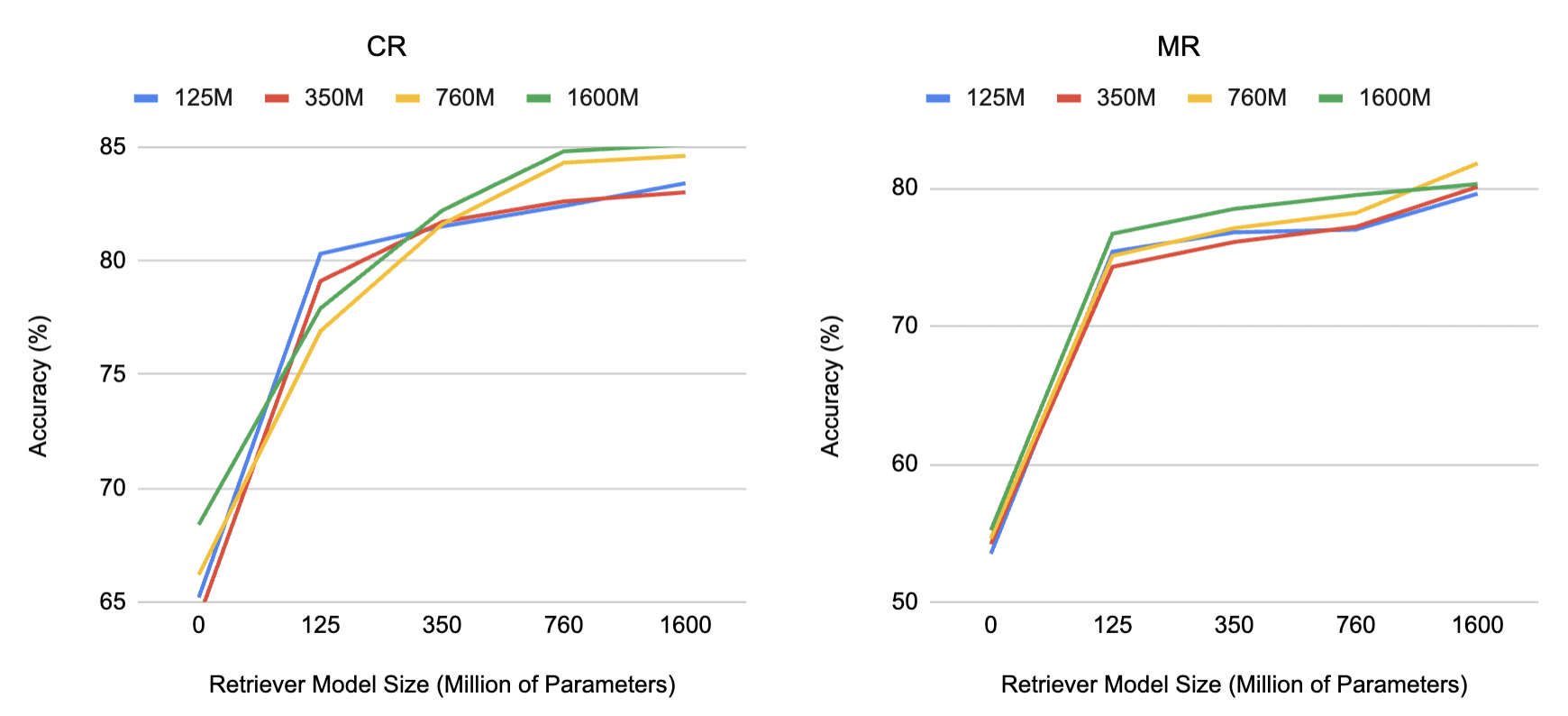}
    \caption{Effect of the retriever model size (GPT-2) on CR and MR. A size of 0 indicates that no retriever is used. Different lines represent different-sized inference models (GPT-2). The benefits of \ourmodel~scale with the retriever model size.
    }
    \label{fig:size}
\end{figure*}

We perform several experiments to understand the contribution of each component of \ourmodel\ and inform our choice of hyperparameters.

\paragraph{Model ablations} \label{aba:pmi_vs_fuzzy}
\ourmodel\ incorporates three features on top of the base LM: $k$NN retrieval and interpolation (\autoref{sec:knnlm}),
fuzzy verbalizers (\autoref{sec:fuzzy-verbalizers}), and PMI scoring (\autoref{sec:full-model}).
\autoref{tbl:ablation} shows the results of ablation experiments analyzing the contribution of each component. 

First, we find that adding kNN to LM gives trivial improvement (+0.4\%),
but much greater once we add fuzzy verbalizers on top of them (+10.3\%), exceeding the contribution of the two components independently (with fuzzy verbalizers alone at +7.2\%).
This supports the argument that fuzzy verbalizers allow the model to make better use of the sparse support of the $k$NN distribution.
Indeed, we find that across all tasks, an output label receives nonzero probability under the $k$NN distribution in $k$NN-LM only 45.8\% of the time. With fuzzy verbalizers, this increases to 78\%.

Second, we find that for the base LM, fuzzy verbalizers bring gains (+7.2\%) similar to PMI scoring (+8.8\%), but the gains are only partially additive when combining the two techniques (+10.9\%).
This suggests that by incorporating more varied surface forms into the score for each label, fuzzy verbalizers may partially --- but not completely --- mitigate the surface form competition problem which PMI scoring was designed to tackle \citep{holtzman-etal-2021-surface}.
The effect of PMI scoring is increased, however, when we use fuzzy verbalizers and $k$NN retrieval together (+13.4\% for the full model versus +10.3\% for $k$NN with fuzzy verbalizers), suggesting that the $k$NN distribution might suffer from more surface form competition problems than the base LM distribution.

\paragraph{$k$NN retrieval hyperparameters}\label{sec:neighbor}
\autoref{fig:neighbor} shows how the number of retrieved nearest neighbors ($k$) and softmax temperature affect model performance on three datasets.
In most cases, performance monotonically improves with the number of neighbors when $k$ is smaller than 1024 and deteriorates after that.
When $k < 256$, a temperature of 1 performs best, while flattening the distribution is useful when retrieving more neighbors. 
Overall, using 1024 neighbors and a temperature value of 3 performs consistently well across the tasks we test.  



\paragraph{Retrieval model size and inference model size} 
\autoref{fig:size} shows how
performance varies with the size of the retriever and inference models on three tasks.
We observe substantial gains as the size of the retriever increases, which hold regardless of inference model size.


It should be noted that a larger retriever leads to a larger datastore and slower retrieval: increasing the retriever size from 125M to 1600M parameters doubles the memory footprint of the datastore, which scales with the size of the retriever model's output embeddings. These computational tradeoffs may inform the retriever size best suited for a particular application.

\section{Related Work}
\paragraph{Retrieval-augmented LMs} Several studies propose the use of retrieval mechanisms with external datastores to improve language modeling performance \citep{Khandelwal2020Generalization} and open-domain question answering \citep{https://doi.org/10.48550/arxiv.2007.01282, rag2020}. Other work incorporates retrieval directly into the LM pretraining process \citep{guu2020retrieval,borgeaud2021improving}. \citet{khandelwal2021nearest} applies nearest neighbor retrieval to conditional sequence generation to improve the quality of machine translation systems. Our work is the first to show that retrieval augmentation, introduced at test time, improves the zero-shot inference of language models on a variety of end tasks. 

\paragraph{Zero-shot and few-shot inference} \citet{DBLP:journals/corr/abs-2005-14165} demonstrate that large LMs  can perform zero-shot (given only a prompt) and few-shot learning (using a concatenation of training examples as demonstrations) without any finetuning. Subsequent work further improves their zero-shot and few-shot abilities with calibration~\cite{holtzman-etal-2021-surface, zhao2021calibrate, min2021noisy}, prompt engineering~\cite{lu2021fantastically, shin-etal-2020-autoprompt} and meta-tuning~\cite{ min2021metaicl, wei2022finetuned, zhong-etal-2021-adapting-language}.
\citet{rubin2021learning} and \citet{liu2021makes} apply retrieval methods to select in-context learning examples that are semantically-similar to a test example for few-shot inference. However, these retrieval methods require access to a large set of labeled data. In contrast, \ourmodel\ only assumes the availability of a heterogeneous unlabeled corpus. 

\section{Conclusions}
We present \ourmodel, a new technique to augment LMs with nearest neighbor retrieval for zero-shot inference on end tasks. \ourmodel~substantially improves zero-shot performance on a wide range of multiple-choice and classification tasks. With a domain- or task-relevant datastore, \ourmodel~enables efficient domain adaptation with no additional training, and its benefits scale with the size of the retrieval model.


\section{Limitations}
Although \ourmodel~significantly improves GPT-2 family models' zero-shot and few-shot performance, it stores high-dimensional vectors for every token in the datastore corpus and performs k-nearest neighbor search for every next token, which incurs significant inference overhead. 
Future work may study compressing the datastore and approximating kNN-search for efficient retrieval.
Careful analysis could also explore datastore curation methods to balance task-relevancy, domain generality, and size. In addition, compared with sentence or document-level retrieval, retrieving tokens at each time step may limit the language model's ability to reason about the retrieved information. Future work may explore if more coarse-grained retrieval and interpolation such as chunks, sentences and documents-level lead to better end task performance. 

Our evaluation of \ourmodel~is limited to GPT-2 family models and eleven end tasks. There are many other tasks and language models for which \ourmodel~can be useful. Future work may study the usefulness of \ourmodel~with larger inference models (i.e: GPT-3) and more diverse tasks. Potentially, large inference models combined with larger retrieval models may result in better zero-shot performance.






\bibliography{anthology,emnlp_2022}
\bibliographystyle{acl_natbib}

\begin{table*}[t]
\centering
\small
\begin{tabularx}{\textwidth}{Xllrr}
\toprule
Test Example & Label & LM Prediction & & \\
\midrule
Too few games could set back PSP launch - Sony exec Signs of a delay, or just managing expectations? The text topic is about & technology & sports & & \\
\midrule
Retrieved Context & Retrieved Value & kNN Prediction & Distance & Corpus \\
\midrule
...References to the game are commonly brought up in other articles about... & software & technology & 33.5 & Wikitext-103 \\
...While it would be easy to point an accusatory finger at Sony and blame them for killing the Dreamcast by overselling the PS2... there's a certain level of intellectual dishonesty in such a stance... [ Sega ]'s poor U.S. support for ... & hardware & technology & 33.7 & Wikitext-103 \\
\bottomrule
\end{tabularx}
\caption{An example from AGN where the kNN gives the correct prediction while LM does not. } \label{case_study}
\end{table*}

\newpage
\appendix
\section{Appendix}
\subsection{Case Study}
We manually check examples where \ourmodel~is better than LM to understand why \ourmodel~improves performance. As shown in \autoref{case_study}, the language model has to know the meaning of the entity "PSP" and "Sony", otherwise it may associate "games" with a sport. kNN is able to match "Sony" in one of the retrieved neighbors, resolving the ambiguity of the word "games".

\subsection{Templates}
\autoref{template} shows the template and verbalizer used for each dataset. 
\begin{table*}[t]
\centering
\small
\begin{tabularx}{\textwidth}{lXX}
\toprule
Dataset & \textcolor{red}{Template} + input & \textbf{Verbalizer} (Fuzzy verbalizer) \\
\midrule
\multirow{2}{*}{ RTE } & \color{black} Time Warner is the world’s largest media and Internet company. \color{red} question: \color{black} Time Warner is the world’s largest company. \color{red} true or false? answer: & \textbf{true} (true, yes, correct, faithful, accurate...) \\
&  & \textbf{false} (false, no, incorrect, wrong, untrue, unfaithful...) \\
\midrule
\multirow{3}{*}{ CB } & \color{red} question: Given that \color{black} What fun to hear Artemis laugh. She’s such a serious child. & \textbf{true} (true, yes, correct, faithful, accurate...) \\
& \color{red} Is  \color{black}  I didn’t know she had a sense of humor. \color{red} true, false, or neither? Answer: & \textbf{false} (false, no, incorrect, wrong, untrue, unfaithful...) \\
& & \textbf{neither} (neither, none, nothing) \\
\midrule
\multirow{10}{*}{ Yahoo } & \color{black} why doesn't an optical mouse work on a glass table? \color{red} topic: & \textbf{society} (society, culture, sociality, group, tribal, organization...)   \\
&  & \textbf{science} (science, math, scientist, knowledge, physics, bioscience...) \\
& & \textbf{health} (health, disease, obesity, medicine, nutrition, well-being...) \\
& & \textbf{education} (education, pedagogy, instruction, school, curriculum, college...) \\
& & \textbf{computer} (computer, internet, network, laptop, progammer, hardware...) \\
& & \textbf{sports} (sports, athletics, sportsman, play, football, basketball...) \\
& & \textbf{business} (business, finance, economics, fund, banking, investment...) \\
& & \textbf{entertainment} (entertainment, music, amusement, game, recreation...) \\
& & \textbf{family} (family, relationships, marriage, household, friendship...) \\
& & \textbf{politics} (politics, government, geopolitics, law, democracy, politician...) \\
\midrule
\multirow{4}{*}{ AGN } & \color{black} Economic growth in Japan slows down as the country experiences. \color{red} topic: & \textbf{politics} (politics, government, geopolitics, law, democracy, politician...) \\
&  & \textbf{sports} (sports, athletics, sportsman, play, football, basketball...) \\
&  & \textbf{business} (business, finance, economics, fund, banking, investment...) \\
& & \textbf{technology} (technology, engineering, science, techinal, science, computer...) \\
\midrule
\multirow{2}{*}{ HYP } & \color{black} Are you sick of Republicans? Or just right-wingers in general? ... \color{red} neutral or partisan? Answer: & \textbf{neutral} (neutral, fair, objective, impartial, disinterested...) \\
&  & \textbf{partisan} (partisan, biased, unfair, prejudiced, unjust...) \\
\midrule
SST-2, CR, & \multirow{2}{*}{ \color{black}Illuminating if overly talky documentary. \color{red} It was } & \textbf{great} (great, good, gorgeous, legendary, perfect, phenomenal...) \\
MR, RT & & terrible (terrible, plain, poor, hideous, upset, awful...) \\

\bottomrule
\end{tabularx}
\caption{The \textcolor{red}{template} and the example (colored black) used for each dataset. We also include the \textbf{standard verbalizer} and (a sample of tokens used in the fuzzy verbalizer).}
\label{template}
\end{table*}

\end{document}